\documentclass[10pt,journal,letterpaper]{IEEEtran}
\usepackage[T1]{fontenc}
\usepackage[utf8]{inputenc}
\usepackage{amsmath,amssymb,graphicx,booktabs,tabularx,array}
\usepackage{microtype}
\usepackage{cite}
\usepackage{placeins,balance}
\newcolumntype{Y}{>{\raggedright\arraybackslash}X}
\newcommand{\affmark}[1]{\textsuperscript{#1}}
\title{AERIN: Aerial Instance Navigation with a Visual World Model}
\author{Zhenzhen Jiao\affmark{1}, %
Rui Tian\affmark{1,2}, %
and Zhineng Chen\affmark{1,3}%
\thanks{\affmark{1}Shenzhen Lingtian Zhixing Technology Co., Ltd., Shenzhen, China.}%
\thanks{\affmark{2}College of Computer Science, Beijing University of Technology, Beijing 100124, China.}%
\thanks{\affmark{3}Institute of Trustworthy Embodied AI, College of Intelligent Robotics and Advanced Manufacturing, Fudan University, Shanghai 200433, China.}}
\begin{document}
\maketitle
\begin{abstract}

Searching from high altitude and navigating to a target instance specified by a reference image is a common application for UAVs. In such tasks, high-altitude observations provide broad ground coverage for efficient search, but targets occupy fewer pixels, making instance verification and action selection more difficult. We present AERIN, a visual world model that uses action-conditioned prediction to guide the UAV toward the target while adjusting its viewpoint to obtain higher-quality observations for subsequent instance verification. High-altitude navigation requires fine-grained target details and scene context relevant to visibility, but repeatedly predicting dense visual representations of this information is computationally expensive. AERIN addresses this challenge with a compact, fixed-size, target-conditioned latent state that separately represents both for efficient action evaluation. Experiments demonstrate that AERIN outperforms the evaluated baselines in target approach and observation quality, while requiring substantially less decision time than similar world models.

\end{abstract}

\begin{figure*}[t]
\centering
\includegraphics[width=\textwidth]{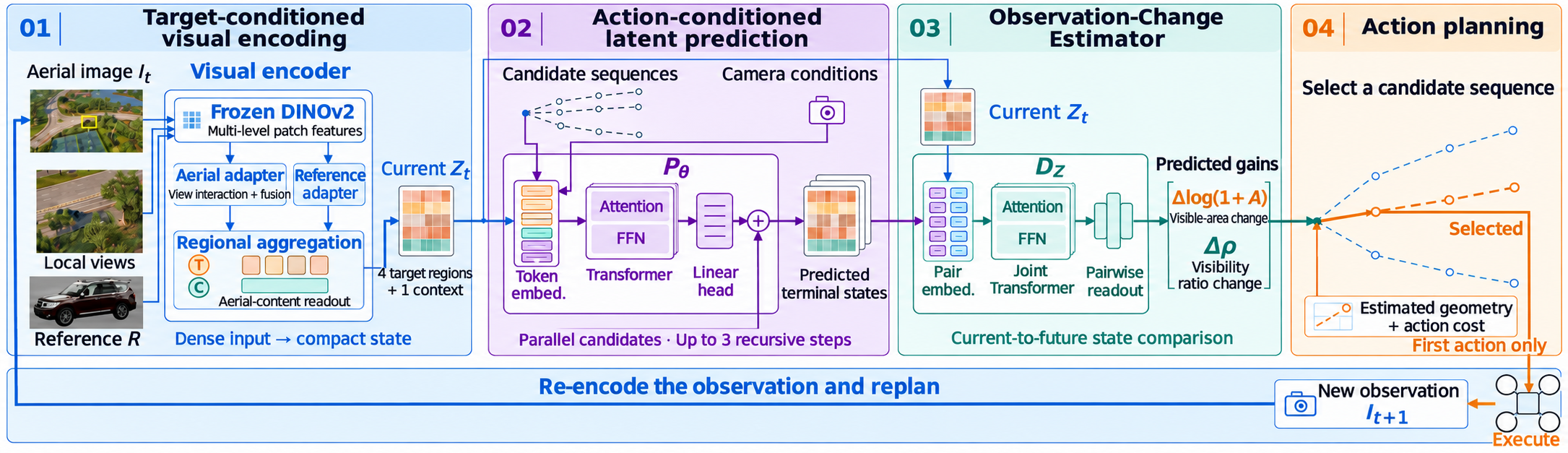}
\caption{Overview of AERIN. Reference-conditioned visual features are aggregated into a compact state for action-conditioned prediction. Predicted observation gains guide action selection, followed by execution and replanning from the new observation.}
\label{fig:1}
\end{figure*}

\section{Introduction}

Instance ImageGoal Navigation (IIN) requires an agent to navigate to a specific object instance depicted in a reference image \cite{ref1}. Existing UAV-based IIN work \cite{ref2} has investigated near-ground navigation in indoor environments, where exploration and collision avoidance are important challenges. In this work, we consider Aerial Instance ImageGoal Navigation (AIIN) in high-altitude patrol settings, where predefined patrol routes can organize regional coverage and broad aerial observations help locate potential targets efficiently. However, targets often occupy few pixels in the high-altitude aerial images and may be partially occluded, providing insufficient evidence for reliable instance verification. Therefore, observation improvement while approaching becomes a key issue in such task, i.e., navigate the UAV toward the target while selecting viewpoints that reveal useful visual details.

Guiding such an approach requires anticipating how UAV motion changes the available visual evidence. Moving closer or descending may enlarge the target in the image, but can also increase its occlusion by surrounding structures, such as a tree canopy. Actions must therefore be evaluated according to their effects on both target detail and visibility. A useful viewpoint may require several moves to reach, even when the first move provides little immediate observation gain. An action-conditioned world model supports such lookahead by recursively predicting how the visual state changes under successive actions, allowing candidate sequences to be evaluated before execution. However, applying world models to AIIN places particular demands on the predictive representation.

First, high-altitude views cover larger areas, facilitating target search, but also contain substantial amounts of task-irrelevant background content, while targets may occupy relatively few informative pixels. Moreover, background changes during flight are often more pronounced than those encountered in near-ground and indoor environments.

Second, in near-ground IIN settings, user-provided reference images are often captured from near-horizontal viewpoints similar to those encountered during navigation. In AIIN, however, UAVs typically observe the target from above, making target details more difficult to identify.

Meeting these observational requirements calls for retaining fine-grained target cues alongside scene context relevant to occlusion. Dense, multiscale feature maps offer one way to preserve both types of information. However, directly using these maps as the predictive state incurs substantial computation, as they must be processed repeatedly across candidate action sequences and rollout steps. This cost can increase planning latency and limit how frequently the UAV replans from new observations. Even with the latest latent-space prediction methods, such as JEPA, this overhead, although reduced, remains substantial. Compressing the state reduces this cost, but may suppress small-target cues, obscure local appearance differences, or discard context relevant to visibility. Therefore, the key challenge is to construct a compact predictive state that preserves the information needed to distinguish the outcomes of different actions while remaining efficient for repeated prediction.

To address this challenge, we present AERIN, a visual world model that guides UAVs toward target while improving observation quality through efficient action-conditioned prediction. Its core is a fixed-size, target-conditioned latent state that allocates representational capacity according to the roles of target details and scene context. Guided by the reference image, the state separately represents local appearance differences within the target region and surrounding information relevant to its visibility. This organization decouples perception resolution from predictive state size: high-resolution visual features inform state construction, while repeated prediction operates on the resulting compact state.

AERIN uses action-conditioned predictions of this state to evaluate the observation outcomes of candidate action sequences. Guided by these predictions, the UAV approaches the target while acquiring clearer, less occluded observations for subsequent instance verification. Experiments show that AERIN improves observation quality and the success rate of acquiring high-quality observations over the compared methods, while requiring substantially less decision time than the evaluated dense-state world models.

\section{Related Work}

\textbf{Visual navigation and active perception.} Image-goal navigation specifies destinations or instances through reference images \cite{ref1,ref3,ref4,ref5}, using feature matching, cross-view learning \cite{ref4,ref5}, or explicit scene representations \cite{ref6,ref7,ref8}. Active perception acquires additional evidence through movement \cite{ref9,ref10,ref11,ref12}, including candidate approach and verification in IEVE \cite{ref13}. Active tracking combines target appearance, observation history, and scene context \cite{ref14,ref15,ref16,ref17}; OA-VAT \cite{ref18} integrates instance association with occlusion recovery. Geometric approaches jointly consider visibility, kinematics, and collision avoidance \cite{ref19,ref20,ref21,ref22,ref23,ref24}. Recent aerial approaches include altitude-aware embodied tracking with AaDWorlds \cite{ref55} and JEPA-based active object detection \cite{ref56}. Our focus is reference-guided target approach and observation improvement after candidate discovery.

\textbf{Visual world models for planning.} Action-conditioned world models support planning through video prediction \cite{ref25}, latent-state prediction \cite{ref26,ref27,ref28}, and policy learning from imagined trajectories \cite{ref29}. Pretrained visual features enable prediction and goal-directed action optimization in feature space \cite{ref30}. In navigation, predicted \cite{ref31} or rendered \cite{ref32} views support trajectory evaluation, while navigation policies provide candidate actions \cite{ref33,ref34}. WoMAP \cite{ref35} and CST-WM \cite{ref36} couple future prediction with task-related readouts for active localization and tracking, respectively. UA-NWM \cite{ref57} uses uncertainty-aware predictions to score trajectories toward a goal view. However, preserving decision-relevant evidence through these predictive pipelines remains a distinct challenge when the target occupies only a small fraction of the image.

\textbf{Compact representations for visual prediction.} Visual computation can be reduced through spatially selective detection \cite{ref37,ref38}, latent bottlenecks \cite{ref39}, and token aggregation \cite{ref40}. Object-centric learning organizes image and video observations into object-level representations \cite{ref41,ref42,ref43}, while object-centric world models capture object dynamics and interactions for prediction \cite{ref44,ref45,ref46} and planning \cite{ref44,ref45}. Causal-JEPA \cite{ref47} combines compact object states with masked latent prediction and efficient planning, while Sparse Imagination \cite{ref48} reduces rollout cost using sparse patch subsets. While object-centric models typically assign slots to different objects, we explore compact, task-relevant representations that allocate capacity across target regions to better capture local details in small-target observation tasks.

\begin{figure*}[t]
\centering
\includegraphics[width=0.90\textwidth]{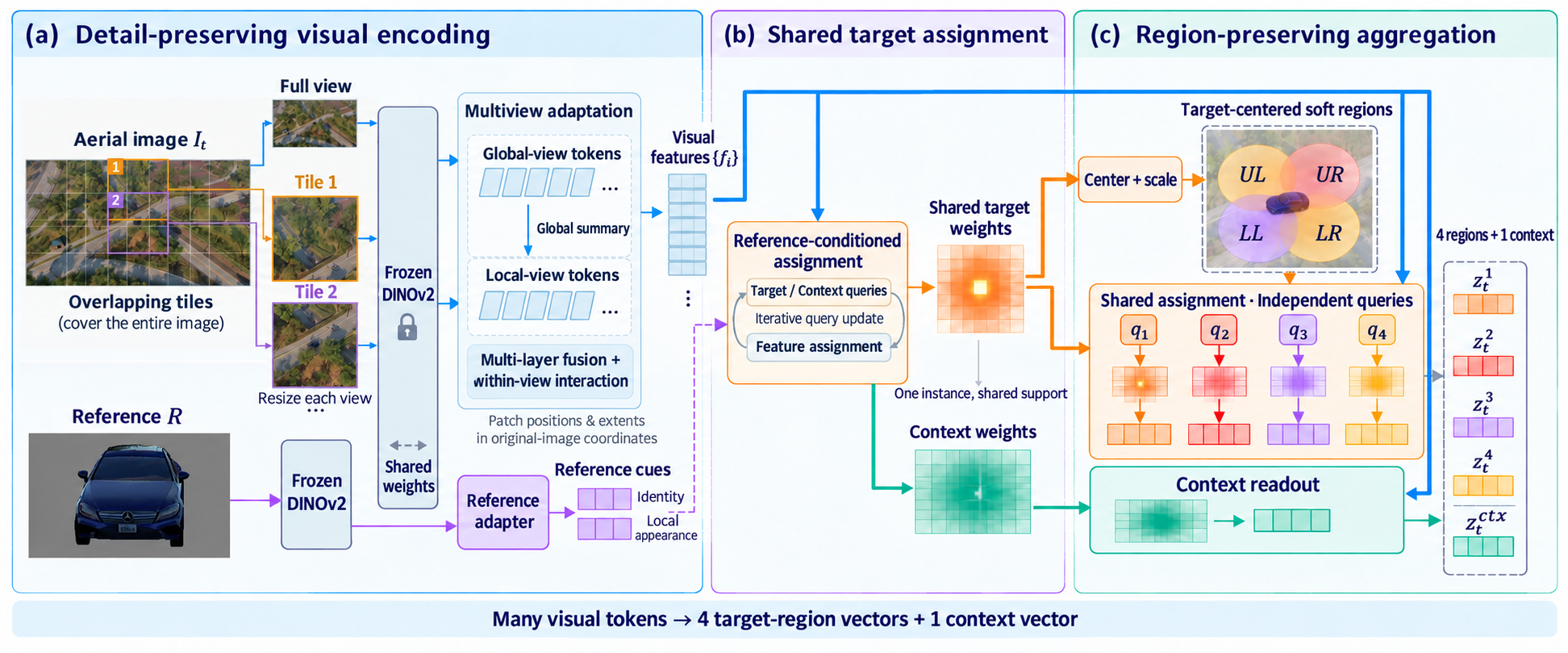}
\caption{Target-conditioned visual encoding. Global and tiled image features are combined with reference cues and aggregated into four target-region slots and one context slot.}
\label{fig:2}
\end{figure*}

\section{Task Definition}
\label{sec:task}

We study Aerial Instance ImageGoal Navigation (AIIN) in high-altitude patrol settings, where a UAV navigates to a specific object instance depicted in a reference image. Given an available map, a geometric coverage planner can generate predefined patrol routes for surveying the search region. Broad aerial views support the discovery of potential targets along these routes, but the observed objects may occupy few pixels or be partially occluded, providing insufficient visual evidence for reliable instance verification. This creates a need to approach candidate objects while actively selecting viewpoints that reveal useful target details.

This patrol-based organization motivates our focus on target approach and active observation. Once a potential target is observed, AERIN uses action-conditioned prediction to guide the UAV's motion, jointly considering progress toward the target and changes in observation quality. The resulting observations can support subsequent instance verification within the overall AIIN workflow: a confirmed match completes the search, whereas a rejected candidate leads to a return to patrol altitude and continued inspection along the planned route.

At each decision step, the agent selects an action consisting of a three-dimensional translation in world coordinates and changes in camera yaw and pitch:
\begin{equation}
a_t = (\Delta x_t, \Delta y_t, \Delta z_t, \Delta \psi_t, \Delta \theta_t).
\label{eq:1}
\end{equation}

The agent selects an action from candidates and a low-level controller executes the corresponding pose command. After execution, the agent receives a new image and updates its decision.

We characterize observation quality using visible target area and visibility ratio. Let $M_t^{\mathrm{vis}}$ denote the observed target mask and $M_t^{\mathrm{full}}$ the corresponding unoccluded target mask, both defined over the same image domain with identical camera parameters. We define:
\begin{equation}
A_t=\left|M_t^{\mathrm{vis}}\right|, \qquad \rho_t= \frac{\left|M_t^{\mathrm{vis}}\right|} {\left|M_t^{\mathrm{full}}\right|}.
\label{eq:2}
\end{equation}

Here, $A_t$ measures the visible target pixel area, while $\rho_t$ quantifies target visibility. Approaching the target generally increases its image scale but may also increase occlusion by surrounding objects. Conversely, changing the viewing direction may improve visibility while temporarily increasing the distance to the target. Active perception must therefore account for target scale, visibility, and progress toward the target.

Let $\mathcal{V}_t$ denote navigation attainment: the UAV is within a predefined distance of the target, the target remains within the field of view, and no collision has occurred up to and including time \emph{t}. High-quality completion additionally requires the target visibility ratio to meet a predefined threshold $\rho_{\min}$:
\begin{equation}
\mathcal{S}_t
= \{\rho_t \geq \rho_{\min}\}
\cap \mathcal{V}_t.
\label{eq:3}
\end{equation}

Completion is evaluated with respect to the reference-specified target using its ground-truth position and visibility annotations. HQC therefore measures successful acquisition of a sufficiently close and visible observation of the correct target. Visible target area is additionally reported as a continuous measure of observation scale. Downstream instance-verification accuracy is not measured by this criterion.

The task terminates successfully when $\mathcal{S}_t$ is first satisfied. Meeting only the distance or visibility requirement does not terminate the task. Execution continues until successful completion or failure. Failure occurs if a collision occurs, the target leaves the field of view entirely, or the decision budget is exhausted without satisfying $\mathcal{S}_t$.

\section{Method}

In AERIN, we organize the visual features into a fixed-size state comprising target regions and scene context, and use this compact state to predict and evaluate observation changes.

\subsection{Overview}

AERIN is organized around a reference-conditioned compact state that connects visual perception, action prediction, and observation planning. Given the current aerial image and the target reference, the encoder aggregates global and local features into four target-region slots and one context slot, forming the state $Z_t$. Conditioned on candidate actions and the corresponding camera conditions, the predictor recursively rolls this state forward. The observation change estimator $D_Z$ compares the current state with each predicted terminal state to estimate changes in visible pixel area and visibility ratio. The planner combines these estimates with geometric progress and action cost to select a sequence, and the UAV executes its first action.

\subsection{Target-conditioned Visual Encoding}

The visual encoder preserves small-target details and scene context for predicting observation changes. We process the entire aerial image at a higher effective resolution by dividing it into overlapping tiles and resizing each tile independently to the DINOv2 \cite{ref49} input resolution. This increases the relative scale of small targets throughout the image, preserving details that may be lost during full-image resizing. The tiles are encoded alongside a resized full-image view using the frozen backbone. Multi-layer patch features are projected to a common dimension and fused by learned weights. An aerial adapter models within-view interactions and injects a global-view summary into local features. Patch positions and extents are mapped to the original image coordinates for cross-view aggregation. The reference image uses the same backbone with a separate adapter to extract global identity and local appearance features.

The resulting multiview representation contains a large number of visual tokens, making repeated action-conditioned prediction costly. Global pooling reduces this cost but may obscure local target changes. We therefore introduce a reference-conditioned aggregator that separates instance selection from regional feature aggregation. Following the iterative aggregation principle of Slot Attention \cite{ref41}, target and context queries update feature assignments using reference-based relevance scores. The center and scale of the target assignment distribution define four overlapping soft regions: upper-left, upper-right, lower-left, and lower-right. Independent region queries aggregate local content under the shared target assignment, while a separate context query collects scene information.

Let $f_i$ denote an aerial feature, $p_i$ its target assignment weight, $g_{ki}$ its spatial weight for region k, and $\alpha_i$ an area compensation factor. Region queries $q_k$ attend to keys $k_i$ derived from feature content and position. The weighted aggregation produces four target-region slots and one context slot, forming the compact state $Z_t$:
\begin{equation}
\begin{aligned}
w_{ki}
&=
\frac{
\alpha_i p_i g_{ki}\exp(\mathbf q_k^\top \mathbf k_i/\sqrt{d_r})
}{
\sum_j \alpha_j p_j g_{kj}\exp(\mathbf q_k^\top \mathbf k_j/\sqrt{d_r})
},\\
z_t^k&=\sum_i w_{ki}f_i,\qquad k=1,\ldots,4,\\
Z_t&=[z_t^1,z_t^2,z_t^3,z_t^4,z_t^{\mathrm{ctx}}].
\end{aligned}
\label{eq:4}
\end{equation}

Here, $d_r$ denotes the dimension of the query and key vectors. Each slot is a 256-dimensional vector, and the five slots form a $5\times256$ state matrix $Z_t$. The target assignment $p_i$ is obtained through reference-conditioned target/context routing. The spatial weights $g_{ki}$ define four overlapping soft regions based on the center and extent of the target assignment distribution. Aggregation weights are normalized over patches independently for each target region, while the context slot uses normalized weights proportional to $\alpha_i$(1 - $p_i$). The area compensation factor $\alpha_i$ accounts for patch coverage and overlap across views in the original image, reducing the influence of repeated sampling. Appendix~\ref{app:A} details the assignment updates, spatial weighting, and area compensation.

Visible target masks supervise target assignments and regional reads. Auxiliary losses on visible pixel area, visibility ratio, and their changes across real observation pairs encourage the state to retain information relevant to action evaluation. Training stages, auxiliary readouts, and loss definitions are provided in Appendix~\ref{app:A}.

High-resolution features are extracted once per actual observation, while all action-sequence rollouts operate on the same fixed-size representation of K = 5 slots. With a fixed number of conditioning tokens, the predictor's pairwise self-attention cost therefore scales with $K^2$ rather than the squared number of input features, $N^2$. This allows the visual encoder to process finer spatial detail without increasing the state size used for repeated prediction.

\subsection{Action-conditioned Latent Prediction}

We learn action-conditioned transitions in the compact state space, following the representation-space prediction principle of joint-embedding predictive architectures (JEPA) \cite{ref50}. The frozen visual encoder provides prediction targets from actual future observations.

The predictor estimates the next state from $Z_t$, action $a_t$ (3D translation and changes in camera yaw and pitch), and $c_t$ (current altitude and camera orientation). The five state slots are linearly projected, and $a_t$ and $c_t$ are each encoded into a conditioning token through an MLP. A Transformer jointly processes the seven tokens with learnable role embeddings.

The five state-token outputs are passed through LayerNorm and a linear projection to predict a residual update $F_\theta$. We add this update to the current state to obtain the next-state prediction:
\begin{equation}
\begin{aligned}
\widehat Z_{t+1}&=P_\theta(Z_t,a_t,c_t)=Z_t+F_\theta(Z_t,a_t,c_t),\\
\widehat Z_{t+h}&=P_\theta(\widehat Z_{t+h-1},a_{t+h-1},c_{t+h-1}),\quad h>1.
\end{aligned}
\label{eq:5}
\end{equation}

Multi-step prediction recursively feeds each predicted state into the next transition, with camera conditions updated according to the candidate poses. This recurrence allows the same transition model to predict terminal states for candidate action sequences of different lengths, supporting the multi-step lookahead used by the planner in Sec.~\ref{sec:planning}.

We train the predictor on one- and two-step rollouts using recorded action sequences. The second step takes the predicted first-step state as input, exposing the model to recursive prediction during training. At both horizons, predicted states are matched to the frozen encodings of the corresponding real observations. We compute mean absolute errors separately over the four target slots and the context slot, yielding the training objective:
\begin{equation}
\begin{aligned}
\mathcal L_P={}&\sum_{h\in\{1,2\}}\lambda_h\mathbb E\Bigg[
\frac{1}{4d}\sum_{k=1}^4\left\|\widehat z_{t+h}^{k}-\operatorname{sg}(z_{t+h}^{k})\right\|_1\\
&+\frac{\lambda_c}{d}\left\|\widehat z_{t+h}^{\mathrm{ctx}}-\operatorname{sg}(z_{t+h}^{\mathrm{ctx}})\right\|_1\Bigg].
\end{aligned}
\label{eq:6}
\end{equation}

Here, d denotes the slot dimension, $\lambda_h$ weights the prediction horizons, and $\lambda_c$ controls the relative contribution of the context loss. The expectation is taken over training segments of the corresponding length, and sg denotes stop-gradient on the target states.

\subsection{Observation Change Estimation}
\label{sec:change}

We evaluate each candidate action sequence by its expected improvement in observation quality relative to the current view. An Observation Change Estimator, $D_Z$, compares the current state with the predicted terminal state to estimate changes in visible pixel area and target visibility.

We use the visible target area $A_t$ and visibility ratio $\rho_t$ defined in Sec.~\ref{sec:task} to quantify observation changes. For two real observations at t and $t'$, we define the target change and its estimate as:
\begin{equation}
\begin{aligned}
\delta_{t,t'}&=\begin{bmatrix}\log(1+A_{t'})-\log(1+A_t)\\\rho_{t'}-\rho_t\end{bmatrix},\\
\widehat\delta_{t,t'}&=D_Z(Z_t,Z_{t'}).
\end{aligned}
\label{eq:7}
\end{equation}

The logarithm compresses the numerical range of the area term. Positive components indicate increased visible area and improved target visibility, respectively.

$D_Z$ jointly attends to both states, using region and frame identity embeddings to distinguish slot roles. Its readout combines the two frame representations with their signed and absolute differences, capturing relative observation cues from both target regions and scene context.

We train $D_Z$ on frozen encodings of real observation pairs using bidirectional Smooth L1 regression and a directional loss:
\begin{equation}
\begin{aligned}
\mathcal{L}_D
&= \mathcal{L}_{\mathrm{reg}}
   + \lambda_{\mathrm{dir}}\mathcal{L}_{\mathrm{dir}}, \\
\mathcal{L}_{\mathrm{dir}}
&= \underset{j,\sigma}{\operatorname{mean}}\;
   \mathbb{E}_{\sigma y_j > \tau_j}
   \left[m - \frac{\sigma\widehat{y}_j}{s_j}\right]_+.
\end{aligned}
\label{eq:8}
\end{equation}

Here, $y_j$ and $\widehat y_j$ denote the target and predicted changes in log-transformed visible area or visibility ratio. The sign $\sigma\in\{-1,+1\}$ identifies the direction of change; $s_j$, $\tau_j$, and $m$ denote the label scale, change threshold, and margin, respectively. The regression term fits change magnitudes. The directional term averages losses separately for increases and decreases in each component, then weights the nonempty groups equally. It is zero when all groups are empty.

During planning, $D_Z(Z_t,\widehat Z_{t'})$ estimates observation changes from the current state and each predicted terminal state to score candidate sequences. After execution, $D_Z$ compares the states encoded from the actual observations before and after the action. These estimated changes provide feedback for adjusting the scoring weights of planner $G_a$.

\begin{figure*}[t]
\centering
\includegraphics[width=0.95\textwidth]{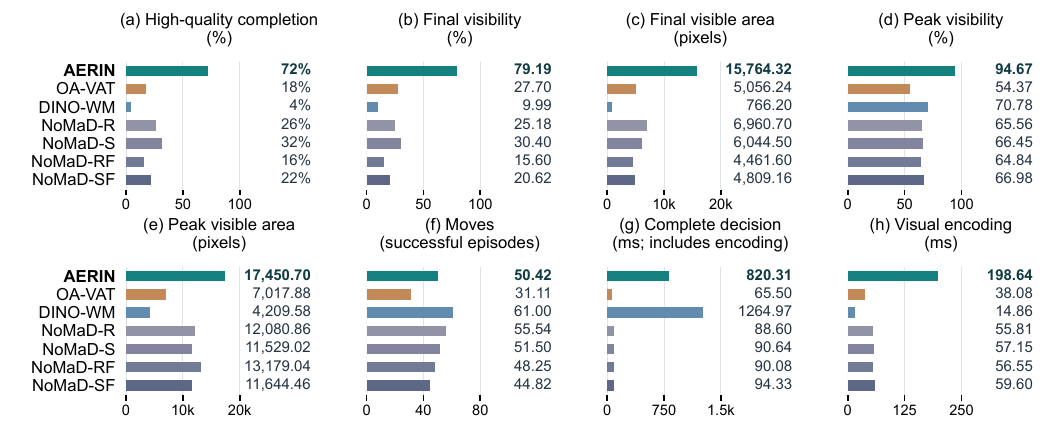}
\caption{Quantitative comparison of AERIN with baseline methods. (a) High-quality completion rate. (b,c) Mean final visibility ratio and visible target area. (d,e) Mean peak visibility ratio and visible target area. (f) Mean number of moves (successful episodes). (g) Complete decision time, including visual encoding. (h) Visual encoding time.}
\label{fig:3}
\end{figure*}

\subsection{Action Planning}
\label{sec:planning}

To make action-sequence search tractable, $G_a$ uses a discrete set of 15 base motion primitives: six axis-aligned translations, eight spatial diagonal translations, and a position-holding action. $G_a$ evaluates sequences of up to three actions that satisfy flight boundary, altitude, and rotation constraints. For each sequence, the predictor rolls out future states, and $D_Z$compares the current and terminal states to estimate the observation gain. The planner combines this gain with geometric progress toward a fixed target anchor, defined as the target's estimated 3D position obtained from the observation. This anchor also guides camera orientation. With scale normalization absorbed into the weights, each sequence is scored as:
\begin{equation}
\begin{aligned}
&J_t(\mathbf a)={}w_A(t)\widehat{\Delta\ell_A}+\eta_t w_\rho^+(t)[\widehat{\Delta\rho}]_+\\
&\quad -w_\rho^-[-\widehat{\Delta\rho}]_++w_d(t)\Delta d+w_h(t)\Delta h-\lambda_{\mathrm{step}}|\mathbf a|.
\end{aligned}
\label{eq:9}
\end{equation}

Here, $\ell _A$ is the log-transformed visible pixel area, $\Delta d$ is the reduction in distance to the anchor, $\Delta h$ is the reduction in excess height above the approach altitude, and |a| is the sequence length. The length penalty favors shorter sequences with comparable gains. The factor $\eta _t$ discounts positive visibility gains according to execution feedback. Details of target localization, action construction and constraints, and scoring-weight updates are provided in Appendix~\ref{app:A4}--\ref{app:A6}.

\begin{figure*}[t]
\centering
\includegraphics[width=0.90\textwidth]{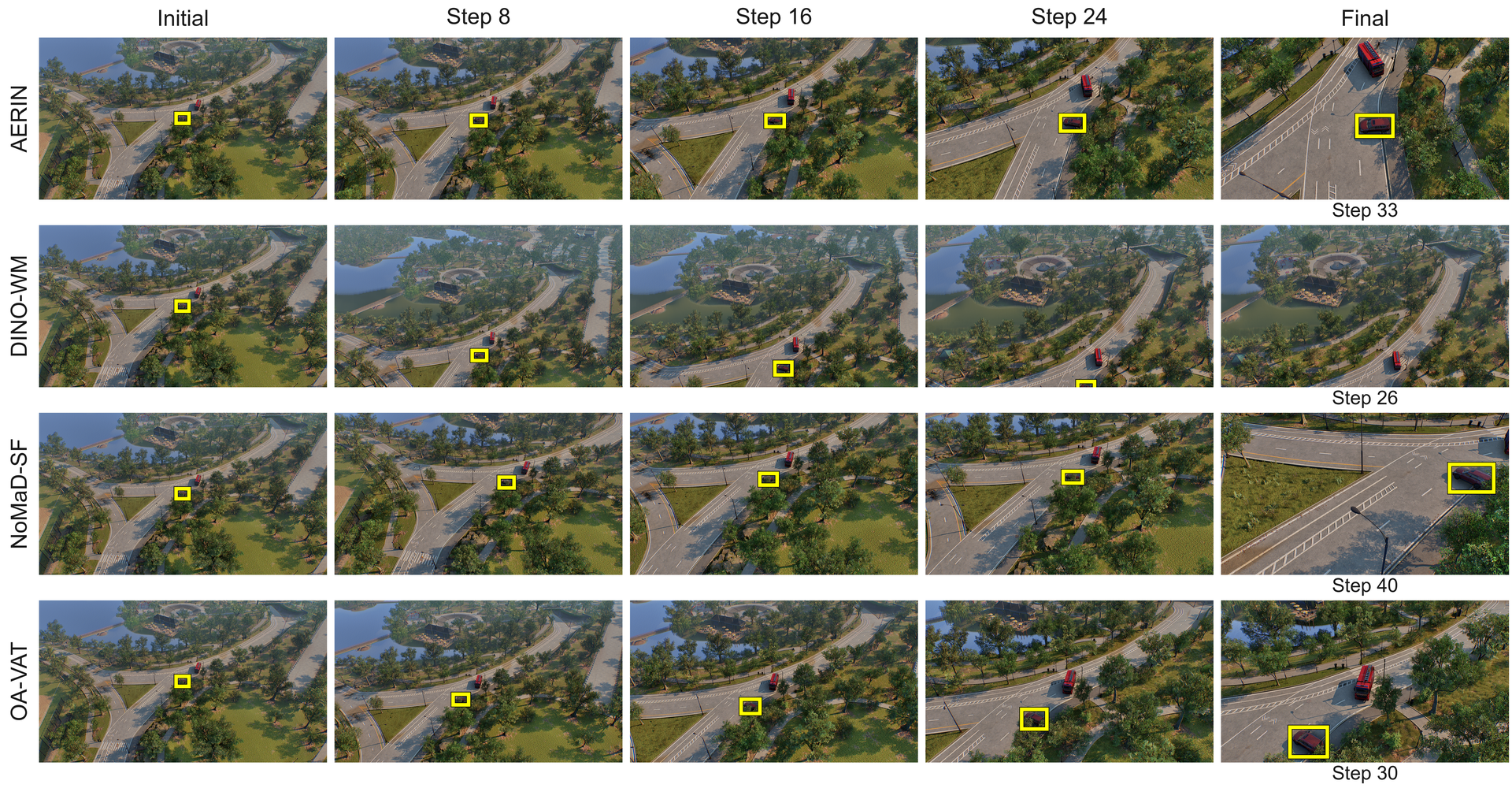}
\caption{First-person observation sequences of different methods in the same scene. Columns show observations at common decision steps, followed by each method's terminal observation with its step and outcome. Yellow boxes mark the visible target extent for visualization.}
\label{fig:4}
\end{figure*}

\section{Experiment}

\subsection{Experimental Setup}
\label{sec:setup}

We evaluate our method against state-of-the-art approaches on a simulation platform developed using Unreal Engine and CARLA \cite{ref51}. The platform supports 3D viewpoint control, occlusion detection, and closed-loop evaluation (Appendix~\ref{app:B}). All training and evaluation are conducted on NVIDIA GeForce RTX 5090 GPUs.

Training, validation, and test configurations are sampled independently by randomly selecting maps and vehicle assets and randomizing vehicle placements and UAV initial poses. Maps and assets may be shared across splits; evaluation uses newly sampled scene configurations.

All methods are evaluated on the same 50 episodes using a common external evaluator, the completion and failure rules in Sec.~\ref{sec:task}, and a budget of 100 actions per episode. Each episode starts at an altitude of 120 m, with one to five vehicles and an initial target visibility ratio randomly sampled at or below 50\%.

High-quality completion rate (HQC) is the percentage of all episodes, including failures, that satisfy Eq.~\eqref{eq:3} within the action budget. Completion requires a UAV--target distance below 35 m and a target visibility ratio of at least 85\% at the same observation, without collision.

Observation quality is measured using simulator annotations after each action. Final visibility ratio and visible target area are taken from the last observation, while peak values are their respective episode maxima; all are averaged over all episodes, including failures. Movement efficiency is measured by the mean number of UAV position changes over each method's successful episodes. We report visual encoding time and complete decision time, with the latter measured from visual input to action selection, including encoding.

\subsection{Comparison with Adapted Baselines}

We compare with OA-VAT \cite{ref18}, DINO-WM \cite{ref30}, and NoMaD \cite{ref34}, representing active visual tracking, visual world modeling, and image-goal navigation, respectively. All methods are evaluated on the same scenes and initial conditions, using the common task-completion criterion, decision budget, and failure rules defined in Secs.~\ref{sec:task} and~\ref{sec:setup}. HQC is computed by the same external evaluator, while each baseline retains its method-specific decision mechanism.

OA-VAT maintains target visibility through instance matching, online tracking, and occlusion recovery. Its released tracking controller disables vertical motion, so we add 3D approach control based on initial visual localization. Its main procedures and the pretrained detection, matching, and tracking modules are all retained. The diffusion recovery network is trained on the same UAV trajectory data used by our method.

DINO-WM predicts action outcomes in dense visual feature space and plans by matching predicted features to a desired goal view. In our setting, the reference image specifies target identity rather than a desired terminal view. We therefore equip DINO-WM with a reference-conditioned Observation Change Estimator trained on the same data and supervision as ours, whose predictions of visible-area and visibility changes support action selection through the shared planner $G_a$.

NoMaD generates 2D navigation waypoints conditioned on a destination image. To support UAV execution, we convert its waypoints into 3D motion and camera commands through a fixed mapping. We evaluate two goal-image configurations: NoMaD-R uses a target reference image consistent with the aerial viewing perspective, while NoMaD-S uses target annotations to extract a fixed square goal crop from the initial aerial image, preserving the target and its surroundings. Both goal images remain fixed throughout the episode. Official pretrained weights are used and we additionally evaluate versions fine-tuned on our dataset, denoted NoMaD-RF and NoMaD-SF, respectively.

Methods requiring a target anchor estimate it from the first frame using their respective visual localization pipelines. Method-specific configurations and training details are provided in Appendix~\ref{app:C}.

\begin{table*}[t]
\centering\small
\caption{Ablation results. (a) Effect of slot aggregation; (b) effect of the planning horizon; (c) effect of predictive state and gain readout; (d) effect of action planning. Bold indicates the best value in each column across all listed configurations. The shared AERIN row uses up to three steps and $Z/D_Z$. The AERIN row serves as the common reference configuration for all four ablation groups.}
\label{tab:ablation}
\setlength{\tabcolsep}{3pt}
\renewcommand{\arraystretch}{1.08}
\begin{tabular*}{\textwidth}{@{\extracolsep{\fill}}lrrrrrrr@{}}
\toprule
Configuration & \shortstack{HQC\\(\%) $\uparrow$} & \shortstack{Final $\rho$\\(\%) $\uparrow$} & \shortstack{Final $A$\\(pixels) $\uparrow$} & \shortstack{Mean peak $\rho$\\(\%) $\uparrow$} & \shortstack{Mean peak $A$\\(pixels) $\uparrow$} & \shortstack{Decision\\(ms) $\downarrow$} & \shortstack{Encoding\\(ms) $\downarrow$} \\
\midrule
(a) Dense-JEPA & 40.00 & 67.82 & 5,438.6 & 89.48 & 6,155.5 & 201,126.36 & 258.08 \\
(a) Generic-5-Slot & 36.00 & 51.63 & 10,511.0 & 89.57 & 14,312.6 & 1,074.07 & 205.08 \\
(b) One-step lookahead & 68.00 & 74.95 & 15,355.5 & 91.40 & 17,072.0 & 399.62 & 211.08 \\
(c) $Z+E$ / quality difference & 56.00 & 69.59 & 14,015.6 & 87.28 & 16,161.2 & 1,091.86 & 201.35 \\
(c) $Z+E$ / $D_{ZE}$ & 64.00 & 72.98 & 15,308.6 & 87.33 & 16,903.5 & 1,015.77 & 206.16 \\
(d) Geometry-only & 12.00 & 24.56 & 8,676.9 & 34.41 & 9,760.2 & \textbf{110.06} & 239.18 \\
\midrule
\textbf{AERIN (all groups)} & \textbf{72.00} & \textbf{79.19} & \textbf{15,764.3} & \textbf{94.67} & \textbf{17,450.7} & 820.31 & \textbf{198.64} \\
\bottomrule
\end{tabular*}
\end{table*}

\subsection{Overall performance}

Fig.~\ref{fig:3} summarizes the quantitative results. AERIN achieves the highest high-quality completion rate, i.e., 72\%. It also achieves the highest mean final visibility ratio of 79.19\% and mean final visible target area of 15,764.32 pixels (Fig.~\ref{fig:3}(b,c)). Its mean peak visibility ratio and visible target area reach 94.67\% and 17,450.70 pixels, respectively, outperforming all compared methods (Fig.~\ref{fig:3}(d,e)). AERIN also achieves competitive movement efficiency, requiring an average of 50.42 UAV position changes to complete the task (Fig.~\ref{fig:3}(f)).

AERIN requires 198.64 ms for visual encoding and 820.31 ms for the complete decision process, including encoding. Its complete decision time is 35.2\% lower than the 1264.97 ms of the adapted DINO-WM baseline (Fig.~\ref{fig:3}(g,h)).

Fig.~\ref{fig:4} presents a qualitative comparison of the four methods in the same scene. AERIN progressively improves target visibility while keeping the target near the image center. NoMaD-SF and OA-VAT also complete the task, but their terminal observations place the target closer to the image boundary. DINO-WM loses the target from view and terminates at step 26.

\subsection{Ablation experiments}

We conduct four groups of ablation studies to examine state representations, prediction horizons, observation-change estimators, and planning strategies. All variants are evaluated on the same 50 episodes, following the task definition and evaluation protocol in Secs.~\ref{sec:task} and~\ref{sec:setup}.

\textbf{Effectiveness of Slot Aggregation.} We compare AERIN with Dense-JEPA and Generic-5-Slot using the same multiview features from the frozen visual encoder. Dense-JEPA bypasses aggregation and feeds these features directly into its JEPA predictor; its observation-change estimator follows the design of $D_Z$, with interfaces adapted to dense features. Both AERIN and Generic-5-Slot aggregate the features into five 256-dimensional reference-conditioned slots and receive target-mask supervision together with the same auxiliary supervision for absolute observation quality and temporal quality changes. These auxiliary objectives supervise observation-level quantities without prescribing roles for individual slots. Generic-5-Slot uses five slots without predefined regional or context roles, whereas AERIN assigns four slots to target regions and one to scene context.

The AERIN and Generic-5-Slot aggregators use the same optimization settings and training schedule. Each representation has a separately trained predictor and observation-change estimator, using common training data. Predictor and estimator training budgets, optimizer settings, and learning-rate schedules are matched across variants within each corresponding stage. All predictors are trained on one- and two-step rollouts, and all change estimators are trained on labeled pairs of real observations as described in Sec.~\ref{sec:change}. All variants use the same planning rules described in Sec.~\ref{sec:planning}.

The results in group (a) of Table~\ref{tab:ablation} show that AERIN achieves higher HQC (72\% versus 40\% for Dense-JEPA and 36\% for Generic-5-Slot), better observation quality, and lower complete decision time than both alternatives. The results support the effectiveness of the proposed task-structured aggregation under shared observation-quality supervision.

\textbf{Effect of Multi-step Lookahead.} We compare one-step planning with a planning horizon of up to three steps. Both variants use identical visual encoder, predictor, and observation-change estimator weights, together with the same planning rules and action budget. Only the maximum candidate sequence length changes. This comparison evaluates the benefit of extending lookahead within the same learned world model and planning framework.

In Table~\ref{tab:ablation}(b), extending the maximum planning horizon from one to three steps increases HQC from 68\% (34/50) to 72\% (36/50), while decision time increases from 399.62 ms to 820.31 ms. This comparison illustrates a completion--computation trade-off on the evaluated episodes.

\textbf{Effect of Predictive State and Gain Readout.} During aggregator training, we construct an auxiliary state \emph{E} from slot-level read statistics and estimates of visible pixel area and visibility ratio. We compare $D_{ZE}$-based gain estimation with gains computed directly from predicted $E$. These two variants share the same encoder and predictor, which recursively propagates both $Z$ and $E$. AERIN uses a separately trained $Z$-only predictor, paired with $D_Z$. All three variants share the visual encoder and planning rules.

The comparison between the two \emph{Z}+\emph{E} variants isolates the gain-readout choice under a shared predictor. The results in group (c) of Table~\ref{tab:ablation} show that AERIN achieves the highest HQC (72\%) and best observation quality, with the lowest complete decision time (820.31 ms).

\textbf{Effectiveness of Action Planning.} We compare $G_a$ with Geometry-only, a one-step greedy controller that approaches the target using geometric distance alone. It uses the same fixed target anchor estimated from the initial observation. Visual encoding is used only for initial target localization; Subsequent decisions are based solely on geometry. At each step, it computes the UAV position resulting from each feasible action and selects the action whose endpoint is closest to the anchor. This controller uses neither future visual-state prediction nor observation-gain assessment.

The results in group (d) of Table~\ref{tab:ablation} show that AERIN achieves an HQC of 72\%, compared with 12\% for Geometry-only. The results demonstrate the benefit of evaluating observation gains alongside geometric approach progress. Geometry-only uses vision only for initial target localization. This one-time encoding cost is amortized over the episode, yielding a lower mean decision time.

Although our closed-loop evaluation uses newly sampled configurations, it is conducted in simulation, where maps and vehicle assets may be shared across data splits. Future work will therefore focus on assessing AERIN's generalization to unseen environments and integrating it with downstream verification and other task-specific modules to enable end-to-end real-world task execution.

\section{Conclusion}

We present AERIN, a visual world model for reference-guided aerial navigation that jointly enables target approach and observation improvement. AERIN employs a compact, target-conditioned latent state to represent target details and scene context, enabling efficient action evaluation while preserving high-resolution visual information. Experiments demonstrate its effectiveness in target approach and visibility improvement, with substantially lower decision time than the baseline world model.

\appendices
\renewcommand{\thesubsection}{\thesection.\arabic{subsection}}
\renewcommand{\thesubsectiondis}{\thesubsection.}
\setcounter{table}{0}
\renewcommand{\thetable}{S\arabic{table}}
\section{State Construction, Training, and Online Planning}
\label{app:A}

AERIN encodes the current aerial observation and the target reference image into a fixed-size state. It predicts how this state will change under candidate action sequences and selects actions based on observation gains, progress toward the target, and action cost. The process repeats after the action is executed and a new observation is acquired.

\subsection{State Construction}
\label{app:A1}

The aggregator takes aerial features, reference features, and patch positions from the visual encoder and produces a state $Z\in\mathbb R^{5\times256}$. Four target-region slots summarize the upper-left, upper-right, lower-left, and lower-right regions of the target, while one context slot summarizes the surrounding scene. Each slot is a 256-dimensional feature vector.

\textbf{Target assignment.} The aggregator uses the reference image to estimate the probability $p_i$ that each aerial patch belongs to the target. A target query and a context query compete for each patch, with reference-matching evidence $m_i$ added to the target score.

\[
(p_i,1-p_i)=\operatorname{softmax}(\ell_{Ti}+m_i,\ell_{Ci}).
\]

Here, $\ell_{Ti}$ and $\ell_{Ci}$ are the query-patch matching scores. Reference evidence is computed from interactions between aerial features and local and global reference features, using an MLP with dimensions $258\rightarrow64\rightarrow1$ and GELU activations. Each query undergoes three rounds of weighted patch reading followed by GRU and residual-MLP updates. The final assignments are recomputed from the updated queries.

\textbf{Overlap compensation.} The full-image view and local crops can cover the same image locations. We account for this overlap with a patch area weight $\alpha_i$. Let $F_i$ be the footprint of patch $i$ in the original image and $c(u)$ the number of views covering pixel $u$. Then

\[
\alpha_i=
\frac{\int_{F_i}1/c(u)\,du}
{\sum_j\int_{F_j}1/c(u)\,du}.
\]

The integral covers valid image content and excludes padding. These weights are used for feature reading during query updates and regional aggregation.

\textbf{Regional aggregation.} We estimate the target center and scale using normalized weights proportional to $\alpha_i p_i$, then define four overlapping soft regions around the center. Let $r_i$ be the patch-center offset normalized by the target scale. The four sign combinations of $\sigma(\pm2r_{ix})\sigma(\pm2r_{iy})$ give the upper-left, upper-right, lower-left, and lower-right memberships. The scale estimate accounts for both the distribution of patch centers and their footprint sizes.

Each region query is initialized from the final target query and a learned region embedding. The four queries then undergo three independent read/update iterations with shared update parameters. Each region normalizes its patch-reading weights separately, as in Eq.~\eqref{eq:4}. The context slot uses normalized weights proportional to $\alpha_i(1-p_i)$. All five slots pool aerial content features, with reference and positional features determining the assignment and reading weights.

\subsection{Aggregator Supervision}
\label{app:A2}

The aggregator receives spatial, single-frame quality, and temporal supervision. Spatial supervision guides each slot toward its assigned region. Quality supervision preserves information about target size and occlusion. Temporal supervision encourages the state to capture observation changes across actions.

Visible target masks undergo the same cropping, resizing, and padding as the RGB images. Averaging the mask within each patch gives its target-assignment label $v_i$. Replacing $p_i$ with $v_i$ in the construction in Sec.~\ref{app:A1} gives the ground-truth regional reading distribution $w_k^*$. The spatial losses are summarized below.

\begin{table}[!htbp]
\centering\footnotesize\setlength{\tabcolsep}{3pt}
\begin{tabularx}{\linewidth}{@{}>{\raggedright\arraybackslash}p{0.28\linewidth}Y r@{}}
\toprule
\textbf{Supervision} & \textbf{Loss} & \textbf{Weight} \\
\midrule
Target/context assignment & Area-weighted soft cross-entropy, normalized by each present class's mass & 1 \\
Reference-matching evidence & Class-balanced binary cross-entropy & 0.25 \\
Regional reading & $\mathrm{KL}(w_k^*\Vert w_k)$ & 1 \\
Regional content & MAE between the slot and frozen aerial features pooled with the ground-truth distribution & 0.02 \\
Region overlap & Squared positive excess of predicted Bhattacharyya overlap over the ground-truth overlap plus 0.05 & 0.02 \\
Region center and spread & Smooth L1 on the mean and standard deviation of the reading distribution & 0.1 each \\
\bottomrule
\end{tabularx}
\end{table}

Center and spread losses use the final reading distributions. Their coordinates are normalized by the componentwise maximum of the ground-truth target scale, mean patch footprint, and $10^{-3}$, with gradients stopped through this normalizer. Regions with zero target mass are excluded from reading, content, center, and spread supervision.

\textbf{Single-frame observation quality.} We supervise the visible target area A and visibility ratio $\rho$ defined in Sec.~\ref{sec:task}. The visible area is estimated directly from the target assignments.

\[
\widetilde A=WH\sum_i\alpha_i p_i.
\]

Visibility is predicted from $Z$ and auxiliary slot statistics. Each slot retains six statistics comprising assignment mass, reading strength, the two-dimensional read center, and the standard deviation along each image axis. For target region $k$, mass and strength are $m_k=\sum_i\alpha_i p_i g_{ki}$ and $u_k=\sum_i w_{ki}p_i$. Centers and standard deviations are computed using $w_{ki}$, and the context uses complementary assignments. Collecting the statistics in $S\in\mathbb R^{5\times6}$ gives

\[
\widetilde\rho=
\sigma\!\left(\mathrm{MLP}\!\left(\mathrm{vec}[\mathrm{LN}(Z),S]\right)\right).
\]

The MLP has dimensions $1310\rightarrow128\rightarrow1$ with GELU activations. Area supervision uses Smooth L1 in $\log(1+A)$ space, with transition parameter 0.5 and weight 0.25. Visibility supervision uses transition parameter 0.1 and weight 1. Queries with an absent target receive an area label of zero. Visibility supervision applies to samples with a valid unoccluded target projection. The single-image objective sums the spatial and quality losses.

\textbf{Temporal change.} The same aggregator independently encodes two real observations to supervise changes in $Q=(\log(1+A),\rho)$. Let $\widetilde Q$ contain the corresponding estimates and let $s$ be the componentwise RMS of the changes over training pairs, with a lower bound of $10^{-4}$. The pair objective is

\[
\begin{aligned}
\mathcal L_{\mathrm{pair}}={}&\tfrac12(\mathcal L_{\mathrm{quality}}^t+\mathcal L_{\mathrm{quality}}^{t'})\\
&+0.25\,\mathrm{SL1}_{1}(\Delta\widetilde Q/s,\Delta Q/s)\\
&+0.1\mathcal L_{\mathrm{dir}}.
\end{aligned}
\]

The directional loss follows Eq.~\eqref{eq:8}, with thresholds of 0.02 for log area and 0.01 for visibility, and a margin of 0.05 in normalized units. It balances the nonempty groups corresponding to increases and decreases in each of the two channels. Single-image and pair batches are interleaved, with spatial supervision applied to single-image batches. Area and query-level losses balance target-present and target-absent samples when both are valid.

\textbf{Supervision in the Generic-5-Slot ablation.} Generic-5-Slot uses the shared target-mask, absolute-quality, and temporal-change supervision described in the main paper. The quadrant-defined regional reading, content, overlap, center, and spread losses listed above are not applied to this baseline. Table~\ref{tab:ablation}(a) therefore compares the structured aggregation and its associated supervision with a generic reference-conditioned five-slot alternative.

\subsection{Training Procedure}
\label{app:A3}

\textbf{Visual frontend.} The aerial image is encoded as a full-image view and a regular grid of overlapping local tiles. The configuration is summarized below.

\begin{table}[!htbp]
\centering\footnotesize\setlength{\tabcolsep}{3pt}
\begin{tabularx}{\linewidth}{@{}>{\raggedright\arraybackslash}p{0.29\linewidth}Y@{}}
\toprule
\textbf{Component} & \textbf{Configuration} \\
\midrule
Original aerial image & $1920\times1080$ pixels \\
\addlinespace[2pt]
Local tile size and layout & $512\times512$ pixels; 3 rows and 5 columns \\
\addlinespace[2pt]
Tile stride & 352 pixels horizontally and 284 pixels vertically \\
\addlinespace[2pt]
Tile origins & $x\in\{0,352,704,1056,1408\}$ and $y\in\{0,284,568\}$ \\
\addlinespace[2pt]
Number of aerial views & 15 local tiles and one full-image view \\
\addlinespace[2pt]
Backbone input & $518\times518$ pixels per view, using bicubic resizing with aspect ratio preserved and centered padding \\
\addlinespace[2pt]
Local-tile resizing & $512\times512\rightarrow518\times518$, without padding \\
\addlinespace[2pt]
Full-image resizing & $1920\times1080\rightarrow518\times291$, with 113 pixels of top padding and 114 pixels of bottom padding \\
\addlinespace[2pt]
Padding and normalization & ImageNet mean-color padding, followed by ImageNet mean and standard-deviation normalization \\
\addlinespace[2pt]
Backbone & DINOv2 ViT-B/14; 14-pixel patches and 768-dimensional patch features \\
\addlinespace[2pt]
Feature layers & Transformer blocks with zero-based indices 2, 5, 8, and 11, using normalized intermediate features \\
\addlinespace[2pt]
Per-layer projection & A separate LayerNorm(768) and linear projection from 768 to 256 dimensions for each selected layer \\
\addlinespace[2pt]
Layer fusion & A weighted sum of the four projected feature maps, with four learned scalar weights normalized by softmax; the aerial and reference branches have separate projection and fusion parameters \\
\bottomrule
\end{tabularx}
\end{table}

Tile origins are uniformly spaced between zero and the image dimension minus the tile size along each axis. The last tile reaches the image boundary, giving the fixed $3\times5$ layout above. Padding is excluded from the valid image support used for aggregation.

\textbf{Feature adapters.} The aerial and reference branches use separate feature adapters to transform the four selected DINOv2 feature maps into 256-dimensional patch representations. Each adapter includes the per-layer projections and learned layer fusion described above, adds a learned embedding of normalized two-dimensional patch coordinates, and refines the features through a residual MLP with a hidden width of 1024. The aerial adapter additionally encodes each view's normalized crop position and size, resizing scale, and local-view indicator, and applies one eight-head Transformer layer within each view. A mean-pooled summary of valid full-image patches is normalized, linearly projected, and added to the local-tile features to provide broader scene context.

\textbf{Frontend adaptation.} Before aggregator training, the visual frontend undergoes adaptation for reference-conditioned spatial localization using 623 real-image--reference pairs for training and 670 for validation. SAM2-generated pseudo-masks provide supervision within known image regions. Target-mask mass is normalized into a soft spatial label, and the localization loss uses cross-entropy with prediction normalization restricted to known spatial cells. During adaptation, the DINOv2 backbone remains frozen, while the adapter parameters, including the projection layers and layer-fusion weights, are optimized jointly. The resulting visual frontend is then frozen throughout aggregator training.

We train the aggregator, then freeze it and train the state predictor and change estimator separately.

\begin{table*}[!t]
\centering\small
\begin{tabularx}{\linewidth}{@{}p{0.28\linewidth}c Y p{0.15\linewidth}@{}}
\toprule
\textbf{Stage} & \textbf{Epochs} & \textbf{Learning rate} & \textbf{Batch size} \\
\midrule
Spatial supervision of the aggregator & 12 & $2\times10^{-4}$ & 4 \\
Add single-frame quality supervision & 3 & $5\times10^{-5}$ for inherited parameters, $2\times10^{-4}$ for new quality parameters & 4 \\
Add temporal supervision & 3 & $10^{-5}$ for the aggregation core, $5\times10^{-5}$ for quality parameters & 4 single images, 2 pairs \\
$Z$ state predictor & 6 & Cosine decay from $3\times10^{-4}$ to $10^{-5}$ & 128 \\
Change estimator $D_Z$ & 3 & $3\times10^{-4}$ & 256 \\
\bottomrule
\end{tabularx}
\end{table*}

All stages use AdamW. Aggregator training uses weight decay 0.01, gradient clipping at norm 1, accumulation over two batches, and FP32 computation, with the visual frontend frozen. The predictor and change estimator both use weight decay $10^{-4}$.

The single-image data contain 5,696 training queries and 1,956 validation queries. Temporal training uses 16,756 real observation pairs, each with three reference variants. One variant is selected for each pair in each epoch. Validation contains 4,216 pairs and 12,648 pair-reference records. Each temporal epoch contains 1,424 single-image batches and 8,378 pair batches. The pair loss is multiplied by $1424/8378$ to balance their contributions.

The final aggregator is selected based on validation performance in spatial assignment, absolute observation quality, and temporal change prediction.

\begin{table}[!htbp]
\centering\footnotesize\setlength{\tabcolsep}{3pt}
\begin{tabularx}{\linewidth}{@{}>{\raggedright\arraybackslash}p{0.19\linewidth}Y >{\raggedright\arraybackslash}p{0.31\linewidth}@{}}
\toprule
\textbf{Module} & \textbf{Architecture} & \textbf{Loss settings} \\
\midrule
State predictor  & 4 Transformer layers, 8 heads, hidden width 256, feed-forward width 1024, zero dropout & One-step/two-step weights of 0.8/0.2 and context weight 0.25 in Eq.~\eqref{eq:6} \\
Change estimator  & 2 interaction layers, 4 heads, hidden width 128 & Directional-loss weight 0.1, margin 0.05, and channel thresholds 0.02/0.01 in Eq.~\eqref{eq:8} \\
\bottomrule
\end{tabularx}
\end{table}

The change estimator's regression and direction losses are normalized by the RMS of the training labels.

\subsection{Initial Localization and the 3D Anchor}
\label{app:A4}

Initial localization uses the average of the four target-region reading distributions. Each patch's weight is distributed uniformly over its original-image footprint, and overlapping contributions are summed to form a heatmap. Within a square of half-width 128 pixels around the peak, we retain the eight-connected component containing the peak among pixels whose values are at least half the maximum. Its heat-weighted centroid gives the target location in the image.

A valid localization requires a positive peak and a peak-to-mean ratio greater than 1.2. We back-project the selected pixel into a camera ray and take its first valid intersection with an independently supplied static ground surface. The resulting 3D anchor is

\[
\widehat p=c_0+\lambda^*R_{cw}K^{-1}[u,v,1]^\top.
\]

Here, $c_0$ is the camera position, $K$ is the intrinsic matrix, $R_{cw}$ maps camera directions to world coordinates, and $\lambda^*$ is determined by the ground intersection. Initialization requires a downward-pointing ray, available terrain data, and a valid intersection. Otherwise, initialization is marked as failed. The anchor is estimated from image localization and terrain geometry and remains fixed throughout the episode.

The world coordinate system uses an upward $z$ axis, with zero yaw pointing along $+y$. Images have resolution $1920\times1080$ and a horizontal field of view of $75^\circ$.

\subsection{Candidate Actions and Constraints}
\label{app:A5}

The planner uses the base translations and position-holding action defined in the main text, together with an additional translation toward the fixed anchor. Base translations have length 3 m, with components of magnitude $3/\sqrt3$ m for spatial diagonals. If an anchor-directed move crosses the estimated approach sphere, a shorter move ending on the sphere is also included. The search admits endpoints inside the sphere, and task completion is determined by the evaluation criteria.

The camera is aligned with the UAV's heading and directed downward. To keep the target in view during descent, its yaw and pitch can be adjusted toward the estimated target anchor in small, bounded angular increments.

The search excludes infeasible endpoints and within-sequence pose revisits, retaining shorter sequences when endpoints are equivalent.

\subsection{Scoring, Feedback, and Recovery}
\label{app:A6}

The standard score in Eq.~\eqref{eq:9} combines log-visible-area gain, visibility gain, reduction in distance to the anchor, reduction in the terrain-relative reference height, and a per-step penalty. Observation gains are estimated by comparing the current state with the predicted terminal state.

\textbf{Execution feedback.} The system retains up to five recent executed actions. For each action, the change estimator compares states encoded from the observations before and after execution to obtain the visibility change $\delta_i$. This is compared with the prediction $\widehat\delta_i$ made for that executed first step. The feedback adjusts the weight on target approach and discounts positive visibility gains.

For a history of length $n$ and cumulative change $R_t=\sum_i\delta_i$, the stagnation coefficient is

\[
b_t=\frac n5\left[1-\operatorname{clip}
\left(\frac{R_t}{0.01n},0,1\right)\right].
\]

It is zero when the history is empty and is reset to zero if either the latest change or the cumulative change falls below $-0.01$. Sustained stagnation increases $b_t$, placing more weight on approaching the target.

For historical actions with predicted visibility gain above 0.01, we divide the change estimated after execution by the predicted change, clip each ratio to $[0,1]$, and average the results. This gives the discount $\eta_t$, subject to a lower bound of 0.1. It is set to 1 when no historical action qualifies and applies only to positive visibility gains.

\begin{table}[!htbp]
\centering\footnotesize\setlength{\tabcolsep}{3pt}
\begin{tabularx}{\linewidth}{@{}>{\raggedright\arraybackslash}p{0.48\linewidth}Y@{}}
\toprule
\textbf{Score term} & \textbf{Effective coefficient} \\
\midrule
Log-visible-area gain & $(0.25+0.25b_t)/\log(1.1)$ \\
Positive visibility gain & $\eta_t(4-2b_t)/0.05$ \\
Penalty for negative visibility change & $6/0.05$ \\
Reduction in distance to the anchor & $(0.25+1.75b_t)/3$ \\
Reduction in excess height above the target & $(0.05+0.20b_t)/3$ \\
Penalty per additional action in the sequence & 0.05 \\
\bottomrule
\end{tabularx}
\end{table}

These coefficients include scale normalization and, for positive visibility gain, the feedback discount. Distance is measured to the fixed anchor using Euclidean distance. The height term is $\Delta h=[h(p_t)-20]_+-[h(p_{t+H})-20]_+$, where $h(p)$ is height above the local terrain. The 20 m reference height is used for action scoring: descent toward this level is rewarded, while further descent below it provides no additional height reward. The height term contributes zero when terrain height at the endpoint is unavailable. Feedback updates only the scoring coefficients.

\textbf{Action selection.} The planner applies the following priority order.

1. \textbf{A one-step move that completes the approach.} When the UAV is outside the estimated 35 m boundary, prefer a feasible move ending inside it with predicted visibility change of at least $-0.01$. Among qualifying moves, choose the shortest translation.

2. \textbf{An active recovery move.} If the first rule yields no qualifying move, execute the active recovery move when its predicted visibility change is at least $-0.02$.

3. \textbf{The highest-scoring sequence.} Break ties by shorter horizon, continuation of the previous remaining plan, and a fixed sequence order.

\textbf{Loop recovery.} The planner updates the best observation according to Rule 3 after each executed action. Once at least 12 actions have been executed, it also checks the most recent 12 using Rules 1 and 2.

1. \textbf{Spatial repetition.} Let $p_t$ be the current UAV position and $p_i$the stored endpoints of the last 12 executed actions. All endpoints must lie within 12 m of $p_t$, measured by Euclidean distance. Scanning these endpoints chronologically, each contiguous run within 1.5 m of \emph{pt} counts as one visit; Separate visits require an intervening endpoint outside that radius. The spatial condition is satisfied when there are at least three such visits or every endpoint lies strictly less than 0.5 m from $p_t$.

2. \textbf{Limited observation improvement.} For the same 12 actions, the change estimator compares the real observations immediately before and after each action. Recovery requires the signed sum of the estimated visibility changes to be no greater than 0.01 and the signed sum of the estimated changes in $\log(1+A)$ to be no greater than $\log(1.1)$. The spatial condition and both improvement conditions must hold simultaneously.

3. \textbf{Updating the best observation.} Let $F=(F_A,F_\rho)$ and $R=(R_A,R_\rho)$ be the change-estimator outputs for the best-to-new and new-to-best observation orders, respectively. The area component measures change in $\log(1+A)$. With finite outputs and valid localization, the new observation replaces the stored best when either group below holds; all four conditions in that group must be satisfied:

\[
\begin{aligned}
&\left\{\begin{aligned}
 F_\rho &> 0.01, & R_\rho &< -0.01,\\
 F_A &\geq -\log(1.05), & R_A &\leq \log(1.05);
\end{aligned}\right.\\[3pt]
\text{or}\quad
&\left\{\begin{aligned}
 |F_\rho| &\leq 0.01, & |R_\rho| &\leq 0.01,\\
 F_A &> \log(1.05), & R_A &< -\log(1.05).
\end{aligned}\right.
\end{aligned}
\]

When recovery is triggered, it retraces stored own-position waypoints toward an earlier best observation, using steps no longer than 3 m and the same camera limits. Each stored version of the best observation permits one return. Accepting a better observation cancels the active recovery target. After a return, the action that initiated the unsuccessful departure is excluded for the next three decisions. Recovery relies on the UAV's pose history and model comparisons of real observations.

\subsection{Data Flow of the Table~\ref{tab:ablation}(c) Ablations}
\label{app:A7}

The auxiliary state is $E=(S,Q)$. Here, $S$ contains the 30 slot statistics defined in Sec.~\ref{app:A2}, and $Q=(\log(1+\widetilde A),\widetilde\rho)$ contains two observation-quality estimates, giving 32 dimensions in $E$.

\begin{table*}[!t]
\centering\small
\begin{tabularx}{\linewidth}{@{}p{0.23\linewidth}Y p{0.27\linewidth}@{}}
\toprule
\textbf{Configuration} & \textbf{State prediction and rollout} & \textbf{Planning gain} \\
\midrule
$Z+E$ / quality difference & Takes $(Z,E)$, predicts $(\widehat Z,\widehat E)$, and feeds both outputs into the next step & $\widehat Q_{t+H}-Q_t$ \\
$Z+E$ / $D_{ZE}$ & Uses the same trained predictor and recurrence as the row above & $D_{ZE}((Z_t,E_t),(\widehat Z_{t+H},\widehat E_{t+H}))$ \\
AERIN, $Z$ / $D_Z$ & Uses a separately trained $P_Z(Z_t,a_t,c_t)$, with the predicted $\widehat Z$ fed into the next step & $D_Z(Z_t,\widehat Z_{t+H})$ \\
\bottomrule
\end{tabularx}
\end{table*}

All three configurations share the frozen visual encoder and aggregator. Direct quality differencing uses the two components of $Q$. The estimators $D_Z$ and $D_{ZE}$ are trained separately with independent parameters. Within each estimator, forward and reverse frame comparisons share weights.

\section{A UAV Simulation and Evaluation Platform}
\label{app:B}

\begin{figure}[!htbp]
\centering
\includegraphics[width=\linewidth]{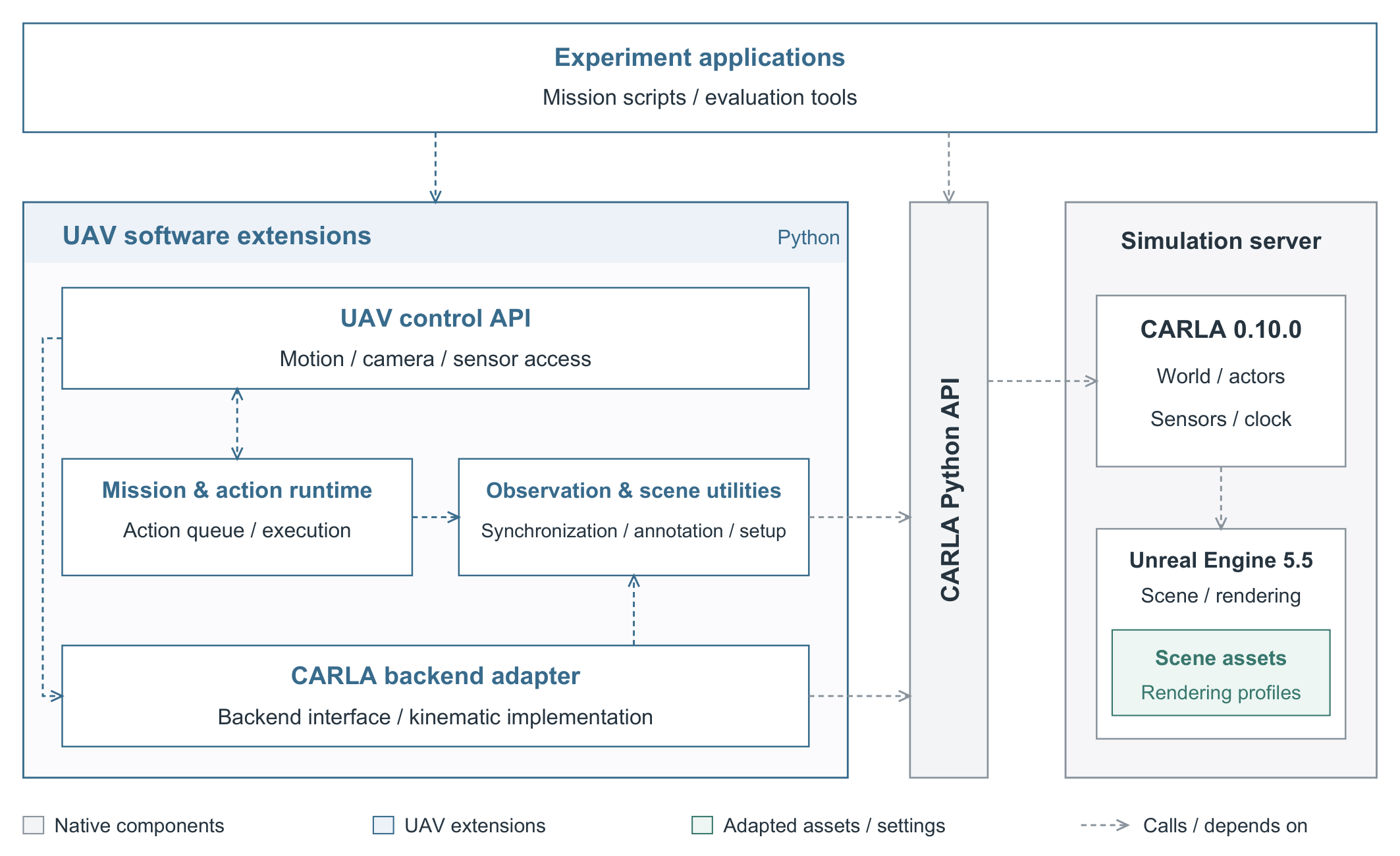}
\caption{Architecture of the UAV simulation and evaluation platform. UAV control, action execution, and observation processing are integrated with the CARLA simulation server.}
\label{fig:5}
\end{figure}

The platform is built on CARLA 0.10.0 and Unreal Engine 5.5 to support aerial observation and closed-loop evaluation (Fig.~\ref{fig:5}). Its architecture integrates UAV and camera control, mission and action execution, and observation synchronization and annotation. These functions interface with the simulation server through the native CARLA API, with a backend adapter providing kinematic UAV motion. Scene assets and rendering settings are further adapted for aerial viewpoints.

\begin{figure}[!htbp]
\centering
\includegraphics[width=\linewidth]{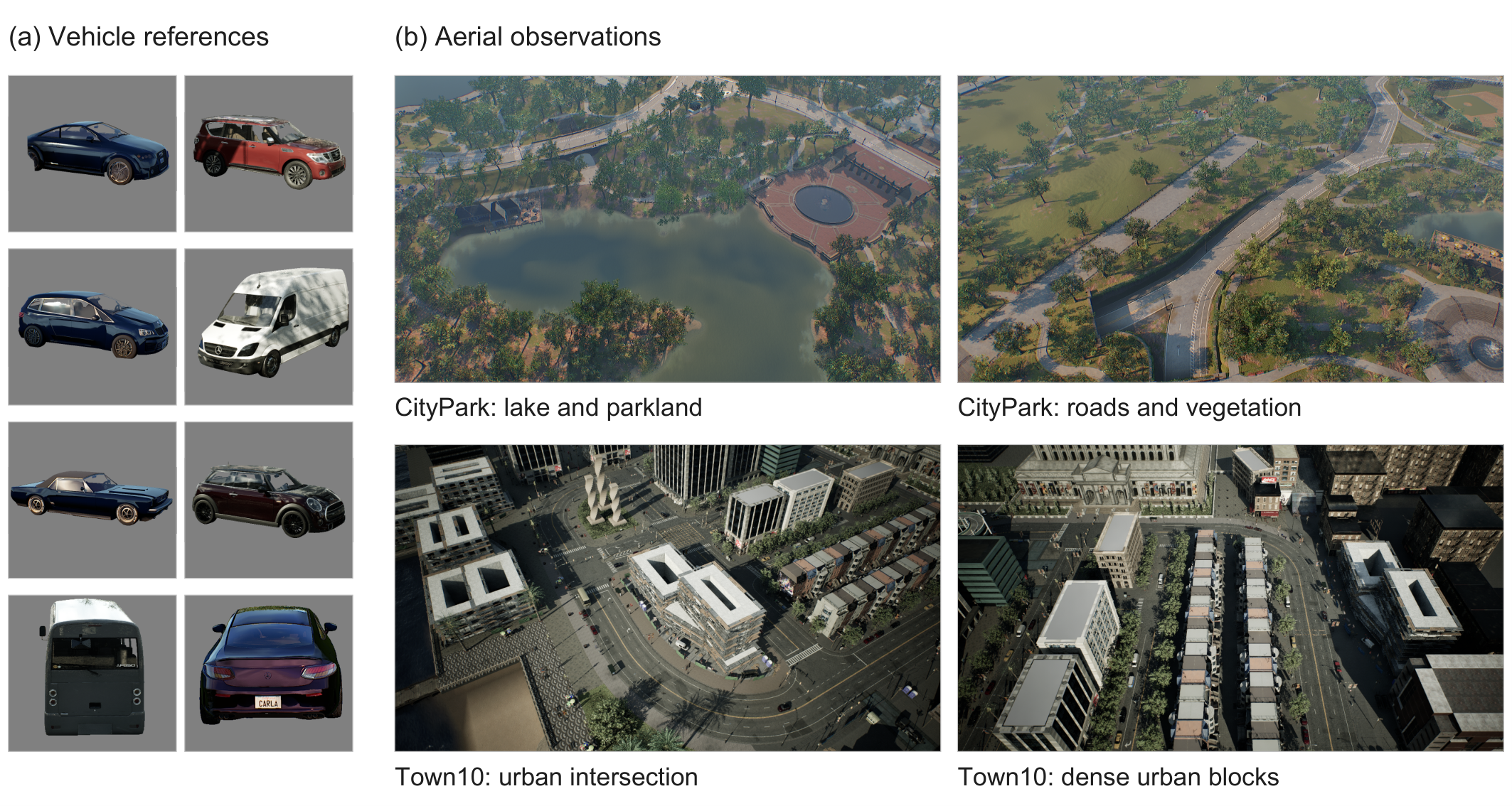}
\caption{Examples of target assets and aerial scenes. (a) Vehicle reference images. (b) Aerial observations from different simulation scenes.}
\label{fig:6}
\end{figure}

The platform provides urban, park, rural, and mining environments with roads, buildings, vegetation, and water, along with vehicle models of varied appearance. By adjusting the UAV position and camera orientation, the platform supports target observations from different viewpoints, with varying target scales, backgrounds, and degrees of occlusion. Figure~\ref{fig:6} shows example vehicle reference images and aerial scenes.

\begin{figure*}[!t]
\centering
\includegraphics[width=0.70\textwidth]{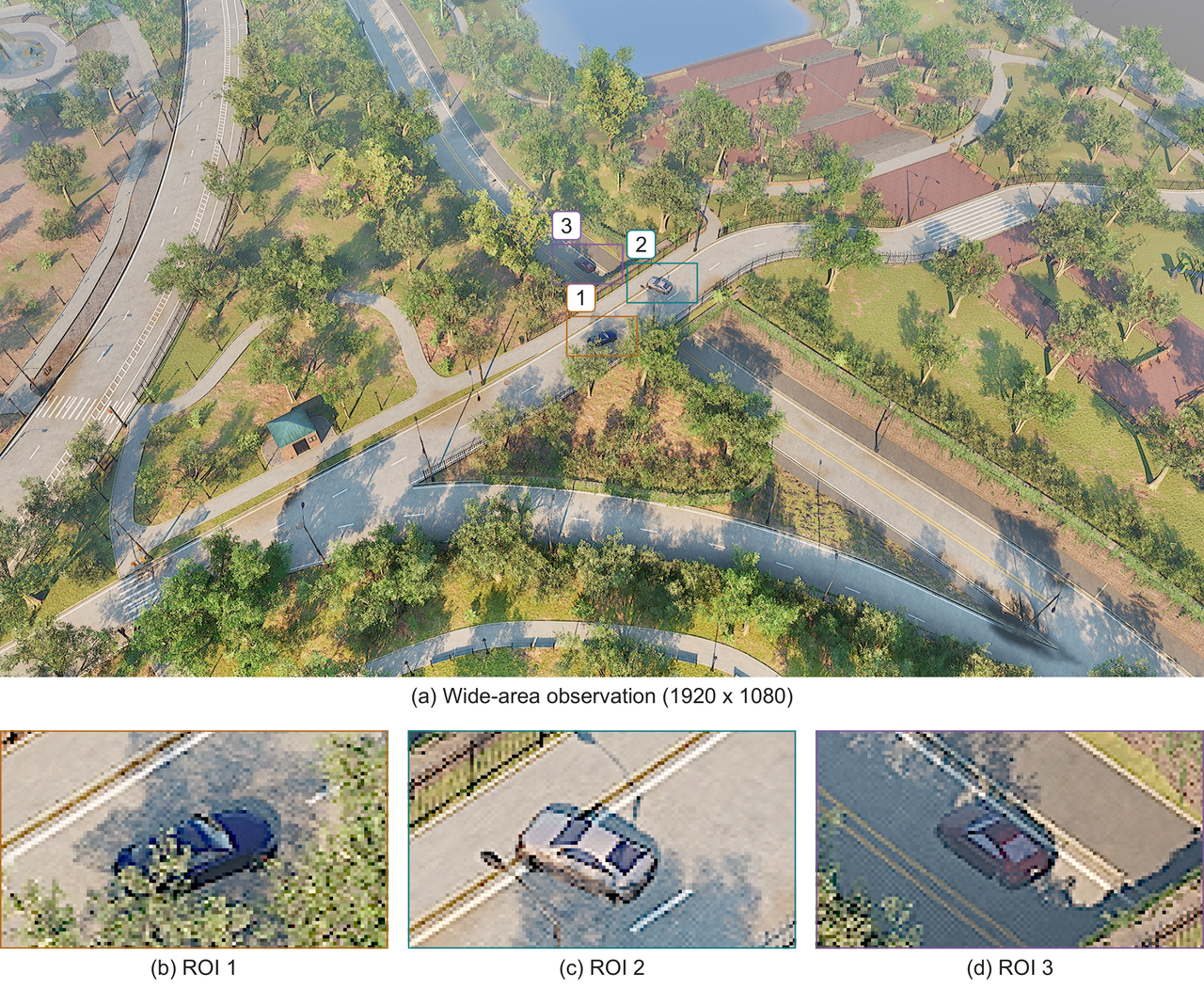}
\caption{Wide-area aerial observation and target details. (a) A $1920\times1080$ aerial observation. (b--d) Enlarged crops of the three marked regions, illustrating target appearance under different occlusion and illumination conditions.}
\label{fig:7}
\end{figure*}

In wide-area aerial images, targets often occupy few pixels, may be partially occluded by vegetation, and can exhibit substantial brightness differences between sunlit and shaded regions (Fig.~\ref{fig:7}). We adjust image resolution, vegetation level of detail (LOD), and post-processing to retain target appearance details, reduce the premature disappearance of distant vegetation, and improve the visibility of targets in shadow, respectively.

\textbf{Image resolution.} We render RGB images at $1920\times1080$ to retain the contours and appearance details of distant targets. RGB and instance-segmentation sensors use the same resolution and imaging geometry, so target localization, visible-area measurements, and observation-quality evaluation share a common pixel correspondence.

\textbf{Vegetation LOD.} At long viewing distances, simplified vegetation can cause low shrubs and other occluders to disappear prematurely, artificially exposing the target. We set the vegetation LOD distance multiplier to 3 to delay these transitions and reduce the resulting visibility artifacts.

\textbf{Exposure and post-processing.} For park and rural scenes, we adjust exposure and local exposure to improve the visibility of vehicles in shaded areas. We also disable motion blur, depth-of-field blur, and vignetting to reduce the blur and brightness attenuation that obscure small-target contours and textures.

\begin{figure*}[!t]
\centering
\includegraphics[width=0.65\textwidth]{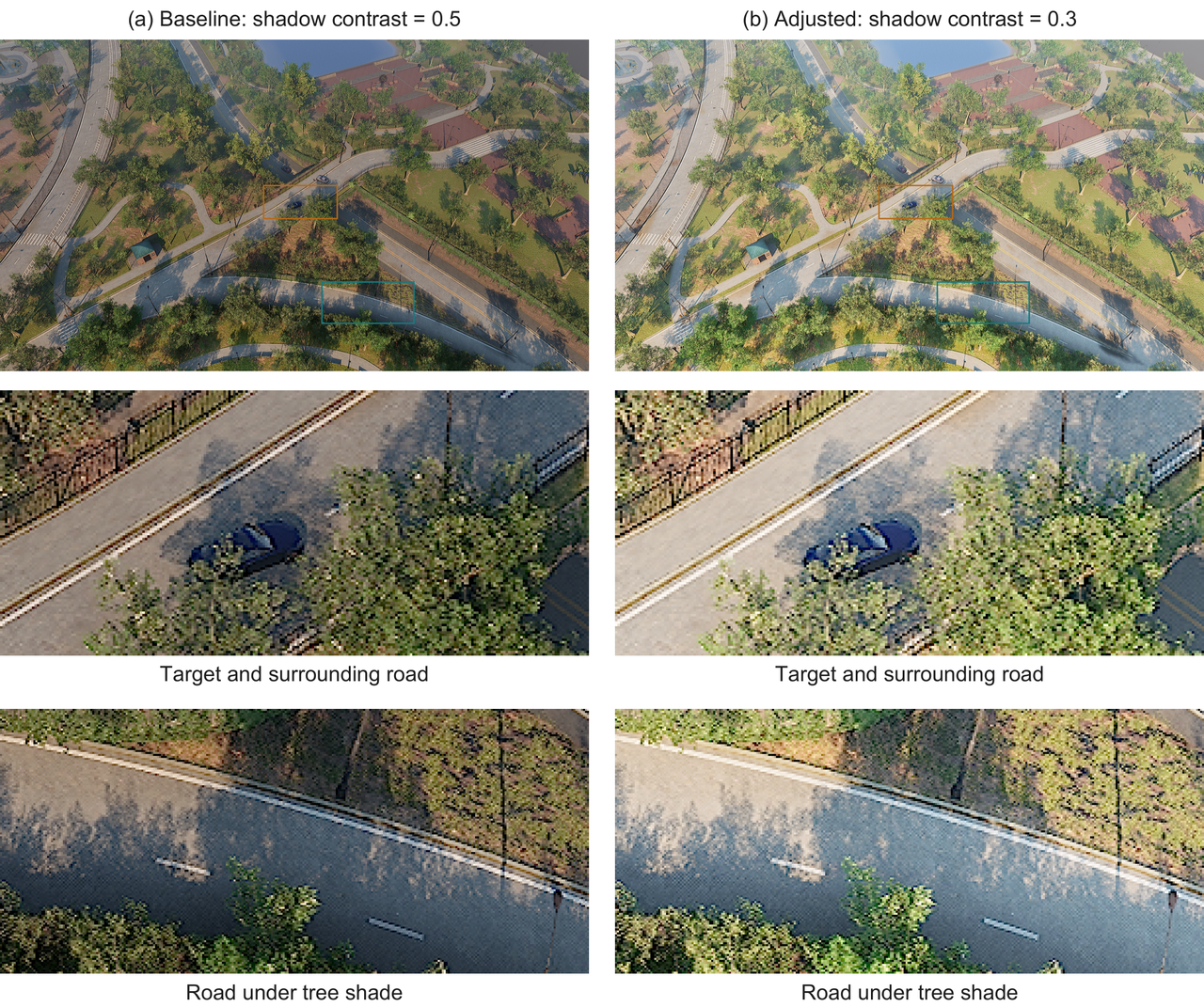}
\caption{Effect of shadow contrast adjustment. The same viewpoint is rendered with shadow contrast set to (a) 0.5 and (b) 0.3. Enlarged crops show the resulting brightness changes around the target and on shaded road surfaces.}
\label{fig:8}
\end{figure*}

All methods use the same initial scenes, observation interface, and task-completion and failure criteria defined in Sec.~\ref{sec:task}. Reference images are also shared, except for the NoMaD goal-image variants in Appendix~\ref{app:C4}. At each step, the interface provides RGB observations, agent pose, and camera parameters; each method uses the inputs required by its design to select an action. The simulator executes the selected action and returns a new observation for the next decision.

An independent evaluator checks the task-completion criterion and failure conditions in Sec.~\ref{sec:task} at initialization and after every executed action, using simulator annotations, including ground-truth target positions and instance masks, to determine task status. These annotations are withheld from online policies. For NoMaD-S/SF, annotations are additionally used to construct a fixed goal crop from the initial image, but only the resulting RGB crop is supplied as this additional input, without target coordinates, annotation boxes, masks, or ground-truth quality values.

\section{Baseline Adaptations and Evaluation Protocol}
\label{app:C}

We adapt OA-VAT, DINO-WM, and NoMaD to AIIN while preserving their core perception and decision mechanisms. The adaptations accommodate instance reference images, approach from high altitude, and UAV control.

\subsection{Comparison Protocol}
\label{app:C1}

All methods share the same scenes, target instances, initial viewpoints, action budget, and evaluation criteria. Online policies may use RGB observations, their own pose, and camera parameters. Ground-truth target coordinates, masks, and observation-quality values are excluded from action scoring. NoMaD-S/SF additionally receive an annotation-assisted goal crop, as described in Sec.~\ref{app:C4}. All evaluated episodes, including those with failed initial localization, are retained in the HQC denominator. Episode outcomes are determined by the common task-completion criterion and failure rules defined in Sec.~\ref{sec:task}. OA-VAT and DINO-WM use the same trajectory data as AERIN's predictor. NoMaD uses the continuous trajectory segments described in Sec.~\ref{app:C4}.

The following table summarizes the method-specific inputs and adaptations. The shared evaluation protocol does not require identical internal representations or control mechanisms. Although several methods use trajectories from the same source, their filtering and training-sample construction differ, as detailed below.

\renewcommand{\thetable}{S\arabic{table}}
\begin{table*}[!t]
\centering\small
\caption{Inputs and adaptations of the evaluated systems. All methods use the common evaluation episodes, action budget, and termination rules.}
\label{tab:supp_protocol}
\setlength{\tabcolsep}{4pt}
\renewcommand{\arraystretch}{1.12}
\begin{tabularx}{\linewidth}{@{}p{0.135\linewidth}Y Y Y@{}}
\toprule
\textbf{Method} & \textbf{Visual and target inputs} & \textbf{Training adaptation} & \textbf{Action selection and execution} \\
\midrule
AERIN & Multiview aerial features and an instance reference; a $5\times256$ state & Frozen DINOv2 backbone; staged frontend adaptation and aggregator, predictor, and gain-estimator training & Predicted observation gains and geometric progress; a visually estimated fixed anchor and bounded camera adjustment \\
\addlinespace[5pt]
OA-VAT & RGB observations, reference matching, and online detection/tracking boxes & Pretrained detection, matching, and tracking modules; trained diffusion recovery planner & Original PID command magnitudes with 3D anchor-directed approach; continuous 10 Hz tracking during execution \\
\addlinespace[5pt]
DINO-WM & Full-image $14\times14$ DINOv2 patch features; reference conditioning in the gain estimator, not the dynamics model & Frozen backbone; separately trained dynamics and reference-conditioned gain estimator & Shared planner $G_a$ using predicted observation gains \\
\addlinespace[5pt]
NoMaD-R/S & Current RGB frame, three preceding frames, and a fixed R or S goal image & Official pretrained checkpoint & Fixed 2D-to-3D waypoint mapping with a $45^\circ$ descent component; no added target-guided camera control \\
\addlinespace[5pt]
NoMaD-RF/SF & Same input formats as R/S & One checkpoint fine-tuned with both R and S goal conditions & Same architecture and control mapping as R/S \\
\bottomrule
\end{tabularx}
\end{table*}

For NoMaD, R denotes an independent top-down target image, whereas S denotes an annotation-assisted crop from the initial aerial image. The S/SF policies receive only the resulting RGB crop as the additional goal input, not its annotation coordinates or masks.

\subsection{OA-VAT}
\label{app:C2}

\textbf{Adaptation.} We retain YOLOE \cite{ref52} detection, DINOv3 ViT-B/16 \cite{ref53} reference matching, ORTrack \cite{ref54}/Kalman tracking, online prototype enhancement, and local and global re-matching. To support approach from high altitude, normal tracking preserves the original PID command magnitudes but directs forward motion toward a 3D target anchor. The anchor is estimated from the initial visual localization, UAV pose, camera parameters, and static terrain elevation, without ground-truth target coordinates.

\textbf{Training.} Detection, matching, and tracking networks retain their pretrained weights. We train only the diffusion recovery planner, including its image encoder. Each sample pairs the current RGB image and target box with the subsequent executed trajectory in the current UAV body frame. Filtering the shared trajectories yields 985 training and 143 validation samples. We train the official diffusion architecture for 100 epochs using MSE noise-prediction loss. Training uses annotated boxes; Online recovery uses detection or tracking boxes.

\textbf{Execution.} The tracker receives 10 Hz video throughout action execution to maintain continuous tracking between decision endpoints. The 65.50 ms decision time reported in Fig.~\ref{fig:3} excludes continuous tracking during action execution; it measures the action-selection stage. It should therefore not be interpreted as the total per-action system runtime including tracking. OA-VAT does not use AERIN's latent predictor, $D_Z$, or $G_a$.

\subsection{DINO-WM}
\label{app:C3}

\textbf{Adaptation.} DINO-WM \cite{ref30} predicts action-conditioned visual features and originally plans toward a goal view. We retain its frozen DINOv2 ViT-B/14 \cite{ref49} backbone and full-image $14\times14$ grid of 768-dimensional patch features, without AERIN's multiview adapter or region aggregator. The dynamics model receives 3D translation and camera rotation, but not the instance reference image.

Our reference image specifies the target instance but does not provide the desired final view required by DINO-WM. We therefore replace goal-image matching with a reference-conditioned observation-gain predictor. We introduce an independent reference-conditioned gain predictor, $D_{\mathrm{DM}}$, that reads the current and predicted dense states to estimate changes in visible pixel area and visibility ratio. It shares no weights with $D_Z$. These estimates support action selection through the common planner $G_a$.

\textbf{Training.} We train the dynamics model using one-step teacher forcing and feature-space MSE on 12,371 unique transitions, with 2,944 for validation. We then freeze it and train $D_{\mathrm{DM}}$ independently on real observation pairs, using 48,252 observation--reference combinations for training and 11,676 for validation. The visual backbone remains frozen throughout.

\subsection{NoMaD}
\label{app:C4}

\textbf{Policy and control.} We retain NoMaD's \cite{ref34} official architecture, which takes the current RGB frame, three preceding frames, and a fixed goal image to predict eight 2D waypoints. We use the local goal-conditioned policy without a topological map or AERIN's localization, gain-estimation, or planning modules.

All four variants use the same fixed UAV control mapping. The forward waypoint component is mapped to a $45^\circ$ descent along the current heading, while the lateral component remains horizontal; each displacement is capped at 3 m. No learned 3D action head or target-guided camera control is added.

\textbf{Goal inputs.} To accommodate NoMaD's destination-image input, we evaluate two goal conditions using the same official pretrained checkpoint. NoMaD-R receives an independent top-down target image with a gray background and no test-scene imagery. NoMaD-S receives a fixed square crop from the initial aerial image, defined by the target's ground-truth amodal box. This crop provides additional annotation-assisted scene information. Neither variant is trained or fine-tuned on our data.

\textbf{Fine-tuned variants.} We fine-tune a single NoMaD model with both R and S goal conditions and evaluate the same checkpoint as NoMaD-RF and NoMaD-SF, respectively. We select trajectory segments that meet NoMaD's training requirements from our existing data and pair each with both goal conditions, yielding 666 training samples and 228 validation samples. Recorded 3D displacements are projected onto the current forward and lateral axes, then converted into normalized 2D waypoints using the official procedure. We update the observation and goal encoders, Transformer layers, and diffusion denoiser, while keeping the temporal-distance head frozen and unused. The architecture and control mapping remain unchanged across all four variants. Fine-tuning retains DDPM noise prediction, goal masking, and EMA, using AdamW for six epochs with a learning rate of $10^{-5}$ and a batch size of 8.

\FloatBarrier
\IEEEtriggercmd{\balance}
\IEEEtriggeratref{25}
\bibliographystyle{IEEEtran}
\bibliography{references}
\end{document}